\documentclass{article}
\usepackage{color}
\usepackage{bm}
\usepackage{stmaryrd}
\usepackage{amsmath} 
\usepackage{listings}
\usepackage{xcolor}
\usepackage{amsthm}
\usepackage{url}
\usepackage{array}

\newcommand\R{{\bf R}}
\newcommand\N{{\bf N}}

\usepackage{tikz}
\usetikzlibrary{shapes, positioning, arrows.meta, decorations.pathreplacing}
\usetikzlibrary{shapes.multipart}

\newtheorem{property}{Property}

\definecolor{codegreen}{rgb}{0,0.6,0}
\definecolor{codegray}{rgb}{0.5,0.5,0.5}
\definecolor{codepurple}{rgb}{0.58,0,0.82}
\definecolor{backcolour}{rgb}{0.95,0.95,0.92}
\lstdefinestyle{mystyle}{
    backgroundcolor=\color{backcolour},   
    commentstyle=\color{codegreen},
    keywordstyle=\color{magenta},
    numberstyle=\tiny\color{codegray},
    stringstyle=\color{codepurple},
    basicstyle=\ttfamily\footnotesize,
    breakatwhitespace=false,         
    breaklines=true,                 
    captionpos=b,                    
    keepspaces=true,                 
    numbers=left,                    
    numbersep=5pt,                  
    showspaces=false,                
    showstringspaces=false,
    showtabs=false,                  
    tabsize=2
}
\title{Transformer for Times Series: an Application to the S\&P500}
\author{Pierre Brugi\`ere\footnote{\text{pbrugiere@ceremade.dauphine.fr, ORCID : 0000-0002-8716-9145}} \  and 
	Gabriel Turinici\footnote{\text{gabriel.turinici@dauphine.fr, https://turinici.com, OrcidID 0000-0003-2713-006X}} \\
	CEREMADE\\
	Universit\'e Paris Dauphine - PSL \\
	Place du Marechal de Lattre de Tassigny \\ 75016 PARIS, FRANCE
}
\date{\today}

\begin{document}
\maketitle
\begin{abstract}
	The transformer models have been extensively used  with good results in a wide area of machine learning applications including Large Language Models 
	and image generation. Here, we inquire on the applicability of this approach to financial time series. We first describe the dataset construction 
	for two prototypical situations: a mean reverting synthetic Ornstein-Uhlenbeck process on one hand and real S\&P500 data on the other hand. 
	Then, we present in detail the proposed Transformer architecture and finally we discuss some encouraging results. For the synthetic data we 
	predict rather accuratly the next move, and for the S\&P500 we get some interesting results related to quadratic variation and volatility prediction. 
\end{abstract}
\section{Objectives and general introduction}

"The transformer models \cite {Vaswani2023}, initiated in 2017, have not yet attained maturity in terms of scientific validation and applicability, despite being used in various contexts of machine learning applications including Large Language Models and generative modeling.

Consistent with this general trend, we inquire here about the applicability of this approach to financial time series.

The outline of this work is as follows: we present the general methodology of our approach in Section \ref{sec:methodology}. We then describe the specific neural network model architecture and the choices we made in Section \ref{sec:architecture}. The results on synthetic Ornstein-Uhlenbeck data and on the S\&P500 are discussed in Section \ref{sec:results}. The final discussion and conclusions can be found in Section \ref{sec:conclusion}."

\section{Methodology} \label{sec:methodology}

\subsection{First notations and time series embedding}

We use the encoder part of a transformer model for time series prediction.
From a time series of unidimensional variables  $(y_1,y_2,\cdots, y_m)$
we build a probabilistic classifier, which from any partial sequence $(\phi(y_{i})$, $\phi(y_{i+1})$, $\ldots$, $\phi(y_{i+l-1}))$ of length $l$ calculates the probabilities for $y_{i+l}$ to belong to each interval of a fixed interval list; these intervals will be referred to as "buckets" hereafter. 
Here, $\phi(.)$ is an embedding function from $\R$ into $\R^d$ which will be discussed later. 
Transformers are highly efficient in large language models, where $y_i$ represents words (or tokens) and  $\phi(y_i)$ denotes their encodings in a $d$-dimensional space.  The value $d$ is called the  dimension of the model and is often set to $512$ or $1024$. Here, to achieve dimension $d$, the undimensional variables $y_i$ are encoded into some new variables $x_i=\phi(y_i)$ with $\phi(y_i)=(y_i, \frac{y_i^2}{2},\cdots, \frac{y_i^d}{d!})$. 
We also incorporate the possibility of including  a positional encoding to the process. Different values for $d$ can be chosen, but in our base case we take $d=\frac{l}{2}$. Details regarding positional encoding are provided later in the paper.

Ideally, the conditional laws 
$\mathcal{L}(Y_{n+l}|(X_{n},X_{n+1},\cdots,X_{n+l-1} ))$ should converge as $n$ increases for the model to perform well.
To investigate this property, we  first utilize simulated data, for which these conditional laws indeed have  a limit. Here, the synthetic dataset is simulated from a  mean reverting Ornstein-Uhlenbeck process, which involves observable variables $y_i$  and hidden state variables $h_i$. We assess the performance of the model by comparing its results to the values derived from $\mathcal{L}(Y_{n+l}| \mathcal{F}_{n+l-1})$. Here, 
$\mathcal{F}_{n+l-1}$ contains all the variables (observable or hidden) up to time $n+l-1$, providing the best possible estimator. 
Hence, in our simulations, $\mathcal{F}_{n+l-1}$ encompasses not only the knowledge of the $(x_i)_{ i \leq n+l-1}$ but also includes the hidden variables $(h_i)_{i\leq n+l-1}$. Subsequently, we test the method on market data with the S\&P500, initially predicting the next day return and then the next day quadratic variation from which avolatilyt prediction could derive.

\subsection{Creating a dataset from a single time series}
We observe a single time series $(y_i)_{i \in \llbracket 1,m \rrbracket}$.
To ease notations, we denote  $[x]_{i}=(x_i,x_{i+1},\cdots , x_{i+l-1})$ a sequence of length $l$. Recall that $x$ is a notation for $\phi(y)$. 

From the observation of $y$ and their embedding $x$, there are different types of datasets of sequences that can be constructed  to train the model:
\begin{enumerate}
\item a single dataset of non-overlapping sequences $[x]_1$ to predict $y_{l+1}$,
$[x]_{l+1}$ to predict
$y_{2l+1}$ and so on
\item a single dataset of overlapping sequences $[x]_1,[x]_2,\ldots , [x]_{m-l}$ 
with the corresponding values to predict $y_{l+1},y_{l+2}, \ldots y_m$

\item $l$ datasets $D_1,D_2..D_{l}$ of non overlapping sequences. Then, classifiers can be calibrated on each of these datasets and combined with an ensemble method. In our case we would have: \\
$D_1=\{[x]_1, [x]_{l+1}, \cdots\}$ \\ 
$D_2=\{[x]_2, [x]_{l+2}, \cdots\}$\\
$\vdots$ \\
$D_{l}=\{[x]_{l}, [x]_{2l}, \cdots\}$.
\item A bootstrap method to build, as in method 3,  several datasets made of sequences  $[x]_i$ where each $[x]_i$ is picked at random from the dataset of all sequences, and then apply an ensemble method to this ensemble of datasets.
\end{enumerate}

Here, we use method 2 and split the dataset of sequences into $\{[x]_1, [x]_2,\cdots [x]_{m_1}\}$ for training and $ \{ [x]_{m_1+1}, [x]_{m_1+2} \cdots [x]_{m-l} \}$ for testing.

\subsection{Ornstein Uhlenbeck process for the $y_i$}

A  time series $(y_1,y_2,\cdots, y_m)$  is generated from some hidden variables $h_i$, in the following way :
 \begin{equation} \left \{ 
 \begin{tabular}{c}
	$h_0 \sim \nu_0$ \\
	$ \forall n \in \N,  h_{n+1} = h_n + \theta (\mu - h_n)dt + \sigma \sqrt{dt} \epsilon_{n+1} \mbox{ with } \epsilon_{n+1} \sim N(0,1)$ \\
	$y_{n+1}= h_{n+1}-h_n $
\end{tabular}
\right  . 
\label{eq:ouprocess}
\end{equation}

We use $\theta >0$ to get a stationary process for the $(y_i)_{i \in \N}$ and take in our simulations $\mu=0$, $dt=1$, $\sigma=1$ and $\theta=1$.

\begin{figure}[htb!]
\centering
\includegraphics[width=\linewidth]{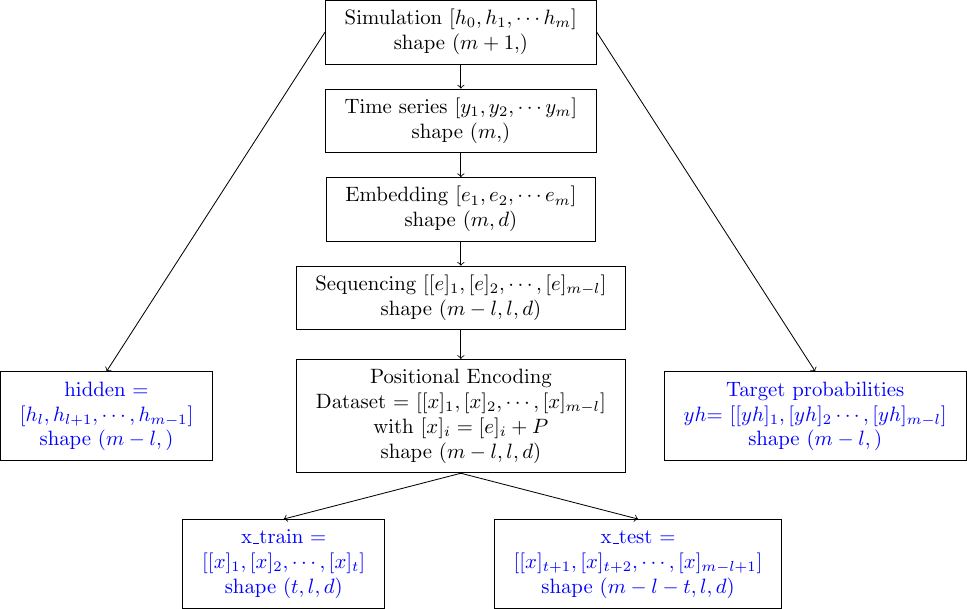}
\caption{Example of variables used in the simulations.}
 \label{fig:variables}
\end{figure}

\begin{figure}[htbp!]
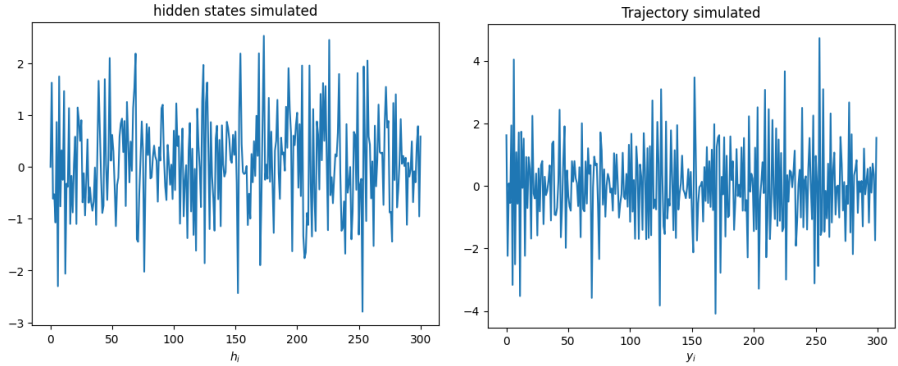

  \centering
  \includegraphics[width=0.49\textwidth]{simulation_h.pdf}
  \includegraphics[width=0.49\textwidth]{simulation_y.pdf}
  \caption{Trajectory of the process \eqref{eq:ouprocess} for the 
  	parameters $dt=1$, $\theta=1$, $\sigma=1$, $\mu=0$.
  	  	{\bf Left:} first 301 hidden values $h_0, ...,h_{300}$.
  	{\bf Right:} first 300 values $y_1, ...,y_{300}$.}
  \label{fig:trajectory}
\end{figure}

As only one trajectory is modeled (see Figure \ref{fig:trajectory}), we choose arbitrarily $h_0=0$ and from there, the hidden values $(h_1,h_2,\cdots h_m)$ are simulated as well as the  $(y_1,y_2,...y_m)$.
Note that, the various  sequences extracted from this time series with method 2 do not have  identical distributions, as each of them is associated to a different initial hidden value $h_0, h_{1},h_{2} \cdots $ etc. This being said, as we can calculate explicitely $\mathcal{L}(Y_{n+l}| \mathcal{F}_{n+l-1})= \mathcal{L}(Y_{n+l}| H_{n+l-1})$ it is not an issue to evaluate the performance of the model by comparing it to the best possible classifier, which is obtained by calculating the probabilities knowing all the variables observed and hidden at the previous time step. The variables used are recapitulated in Figure \ref{fig:variables}.

\subsection{Positional encoding}
As a common practice for transformer models, we enable the model to integrate positional encoding. The positional encoding has the same dimension as the model. So, for a sequence $[x]_i$ of shape $(l,d)$ the positional encoding will be a matrix $P \in \R^{l\times d}$.  The positional encoding could be be learned, but we use here the standard method with sinusoidal functions. For this method, it is better to have an even number for $d$, which will be the case here.\\
For $j \in [0,\frac{d}{2}-1]$ we define $$w_j= \frac{1}{10000^{\frac{2j}{d}}}$$ 
and for $t \in [0,d-1]$, the column vector $p_t$ of $P$ is defined as:
 \begin{equation} \left \{ 
 \begin{tabular}{c}
	$p_{t,2j}=sin(t w_j)$ \\
	$p_{t,2j+1}=cos(t w_j) $
\end{tabular}
\right  . \end{equation}

 This positional encoding presents several interesting properties. 

\begin{property}
\mbox{ }
  For $k \in [0,d-1]$ let $T_k$ be the $2\times 2$  block diagonal matrix such that,   $\forall j \in [0,\frac{d}{2}-1], $ each of its $2\times 2$  diagonal block $M_k(j)$  is defined by: 
   \[ M_k(j)=
\begin{bmatrix}
  cos(kw_j)& sin(kw_j) \\
  -sin(kw_j) & cos(kw_j) \\
\end{bmatrix}
\] then,
\begin{enumerate}
\item  $\forall k \in [0,d-1], T_k$ is an orthonormal matrix. i.e $T_k T^{\perp}_k = T^{\perp}_k T_k =Id_d$
\item $T_{k+t}=T_k T_t$
\item $\forall t \in [0,d-1],  \| p_t\|^2=\frac{d}{2}$
\item  $p_{t+k}=T_kp_t$
\item  for all vector columns $p_t$ and $p_{t+k}$ of $P$ the quantity $\langle p_t, p_{t+k}\rangle$ only depends on $k$ and is maximum for $k=0$.
\item $\exists$ a function $\psi$ such that $P P^{\perp} =[\psi(|i-j|)]$
 \end{enumerate}
\end{property}

\begin{proof} \mbox{ }
\begin{enumerate}
 \item  simple calculation as $cos^2(kw_j)+ sin^2(kw_j)=1$
 \item demonstration block by block using the properties of the trigonometric functions
 \item simple using that $sin^2 + cos^2=1$
\item  demonstration block by block 
  \[ 
     M_k(j)
\begin{pmatrix}
  p_{t,2j} \\
  p_{t,2j+1} \\
\end{pmatrix} 
 = 
\begin{bmatrix}
  cos(kw_j)& sin(kw_j) \\
  -sin(kw_j) & cos(kw_j) \\
\end{bmatrix} 
\begin{pmatrix}
  sin(tw_j)\\
  cos(tw_j) \\
\end{pmatrix} 
=\begin{pmatrix}
  sin((k+t)w_j)\\
  cos((k+t)w_j) \\
\end{pmatrix} 
\] which proves the result.
\item  $\langle p_t, p_{t+k}\rangle = \langle p_{t+k}, p_t\rangle = p_0^{\perp} T^{\perp}_{t+k}T_kp_0= p_0^{\perp}T_kT^{\perp}_tT_tp_0= p_0^{\perp}T_kp_0$ which only depends on $k$ and $\langle p_t, p_{t}\rangle = \frac{d}{2}$  while by Cauchy Schwartz 
$| \langle p_t, p_{t+k}\rangle |= | p_0^{\perp}T_kp_0     |\leq  \| p_0 \|   \|T_kp_0 \|= \| p_0 \|  \| p_0 \|  =  \frac{d}{2}  $ Q.E.D.
\item This result is a consequence of the previous result. 
 \end{enumerate}
\end{proof}

Property 6 implies that the positional encoding enables, by calculating a scalar product between the positional vectors $p_s $ and $p_t$, to get a notion of time elapsed between time $s$ and $t$.

\subsection{The Transformer model for classification}

After defining some buckets $B_1,B_2,B_k$ the objective of the model is, for a sequence $(x_{m+1},x_{m+2}, \cdots , x_{m+l})$, to estimate the probabilities  $(p_1,p_2,\cdots, p_k)$ for $y_{m+l+1}$ (resp $y^2_{m+l+1}$ for quadratic prediction) to belong to  these buckets. In the base case, we take the number of buckets to be equal to $7$ obtaining the list $]-\infty, \alpha_1 ],
]\alpha_1, \alpha_2],\cdots, ]\alpha_6, +\infty[$ with the $\alpha_i$ corresponding to the quantiles of  the observed $y_i$ (resp $y^2_i$) from the training set. Therefore, each bucket contains the same proportion of $y_i$ (resp $y^2_i$), up to the rounding errors, from the training set.

The model used is the encoder part of a Transformer, which is a natural choice when solving a classification problem, and we extend the Encoder with some dense layers at the end for the classification task. We call:

\begin{itemize}
\item $l$ the length of the sequences $(x_1,x_2, \cdots , x_l)$ used for each prediction. In the base case, $l=32$.
\item $d$ the dimension of the model, which is the dimension of each element $x_i$, or equivalently the number of features for each $x_i$. In the base case here, $d=\frac{l}{2}=16$.
\item $k$ the number of buckets used for classification. In the base case here, $k=7$.
\item $p$ the proportion of sequences of the dataset used for training. In the base case here, $p=80\%$ and the training sequences use the first observations  in chronological order.
\end{itemize}

In the learning phase, for parameters estimation, the sequences are processed by batches. The default number for a batch is $32$.  
It is possible to change the batch size in model.fit. Here, in our base case, we take the batch size to be $64$. Below is an example of python code that sets the parameters of the model~:
\begin{lstlisting}[language=Python]
model.fit(
x_train,
yh_train,
validation_split=0.2,
epochs=30,
batch_size=64,      
callbacks=my_callbacks
\end{lstlisting}

The model is built as a sequence of the following blocks :  

\begin{enumerate}
\item  a block to create a) the dataset of the $[x]_i$ of shape (Nb of sequences,$l$,$d$),  with the positional encoding taken into account as an option and b) the dataset of the $y_i$, to be predicted, of shape (Nb of sequences,$l$,$1$). The dataset is split into a training set and a test set and for each $x_i$ we can associate the corresponding $h_i$ and calculate $\mathcal{L}(Y_{i+1}|H_i=h_i)$.  

\item an Encoder block made of a MultiHead Attention block and a Position-wise Feed-Forward block
\item a "Special Purpose" block for class prediction, once the sequences have been "contextualised" by the Encoder.
\end{enumerate}

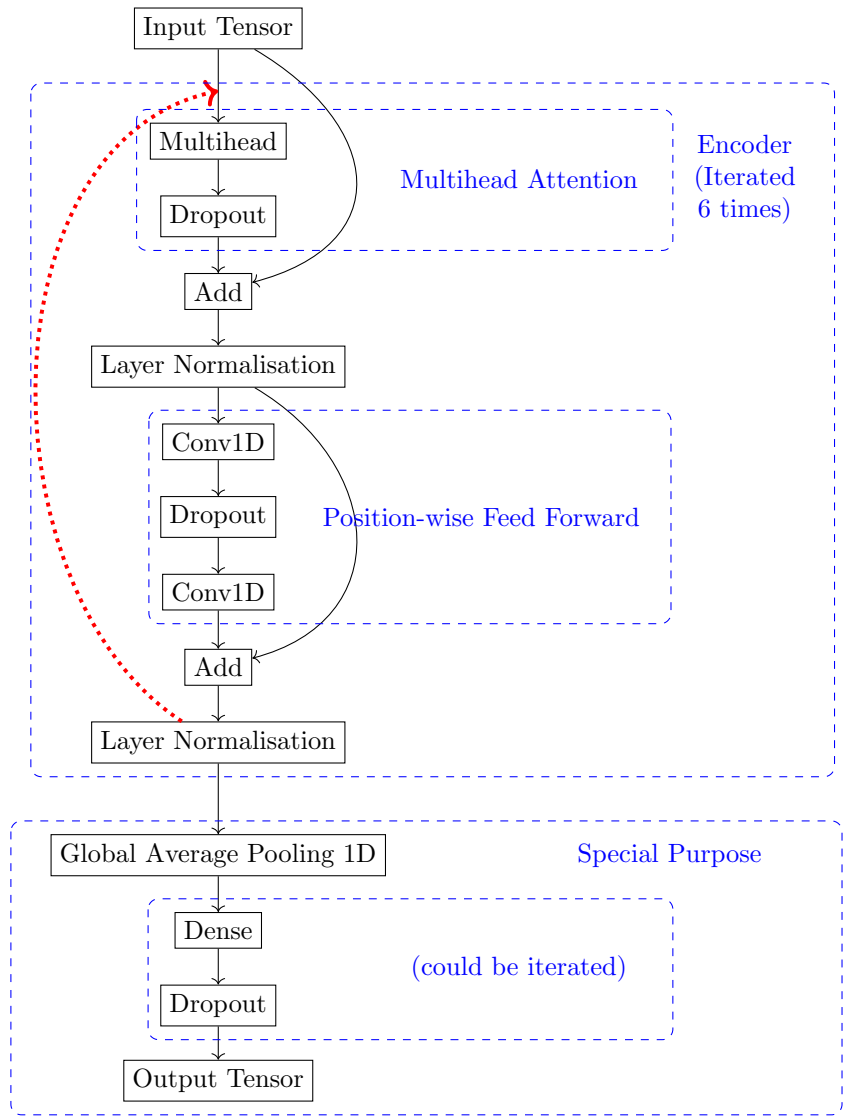
\begin{figure}
  \centering
\caption{Structure of the program.}
\vspace{0.5cm}

\begin{tikzpicture}

  \node[draw, rectangle] (box1) {Input Tensor};
  \node[draw, rectangle] (box3) at (0,-1.5) {Multihead};
  \node[draw, rectangle, below of=box3] (box4) {Dropout};
  \node[draw, rectangle, below of=box4] (box5) {Add };
  \node[draw, rectangle, below of=box5] (box6) {Layer Normalisation};
  \node[draw, rectangle, below of=box6] (box7) {Conv1D};
  \node[draw, rectangle, below of=box7] (box8) {Dropout};
  \node[draw, rectangle, below of=box8] (box9) {Conv1D};
  \node[draw, rectangle, below of=box9] (box10) {Add };
  
  \node[draw, rectangle, below of=box10] (box2) {Layer Normalisation};
  \node[draw, rectangle] (box12) at (0,-11) {Global Average Pooling 1D};
 \node[draw, rectangle, below of=box12] (box13)  {Dense};
\node[draw, rectangle, below of=box13] (box14)  {Dropout};
 \node[draw, rectangle, below of=box14] (box15) {Output Tensor};

 \draw [->] (box1.south) -- (box3.north);
    \draw [->] (box3.south) -- (box4.north);
      \draw [->] (box4.south) -- (box5.north);
        \draw [->] (box5.south) -- (box6.north);
          \draw [->] (box6.south) -- (box7.north);
            \draw [->] (box7.south) -- (box8.north);
              \draw [->] (box8.south) -- (box9.north);
                \draw [->] (box9.south) -- (box10.north);
                   \draw [->] (box10.south) -- (box2.north);
                    \draw [->] (box2.south) -- (box12.north);
              	  \draw [->] (box12.south) -- (box13.north);
	  	    \draw [->] (box13.south) -- (box14.north);
	               \draw [->] (box14.south) -- (box15.north);

  \draw[->, bend right=45] (box1) to [out=60, in=105, distance=2cm] (box5);
  \draw[->, bend right=45] (box6) to [out=60, in=105, distance=2cm] (box10);
    \draw[->, bend right=45, red, dotted, line width=1.5pt] (box2) to [out=60, in=105, distance=3.2cm] ([yshift=12pt]box3.north);


    \draw[dashed, rounded corners, blue] 
    ([shift={(-5pt,5pt)}]box3.north west) rectangle ([shift={(150pt,-5pt)}]box4.south east);
       \draw[dashed, rounded corners, blue] 
    ([shift={(-5pt,5pt)}]box7.north west) rectangle ([shift={(150pt,-5pt)}]box9.south east);
      \draw[dashed, rounded corners, blue] 
    ([shift={(-45pt,15pt)}]box3.north west) rectangle ([shift={(185pt,-5pt)}]box2.south east);
         \draw[dashed, rounded corners, blue] 
    ([shift={(-10pt,5pt)}]box13.north west) rectangle ([shift={(150pt,-5pt)}]box14.south east);
           \draw[dashed, rounded corners, blue] 
    ([shift={(-15pt,5pt)}]box12.north west) rectangle ([shift={(200pt,-5pt)}]box15.south east);

  \node[ text=blue] (box16) at (4,-2) {Multihead Attention};

   \node[ text=blue] (box17) at (3.5,-6.5) {Position-wise Feed Forward};
  \node[ text=blue] (box17) at (4,-12.5) { (could be iterated) };
  \node[ text=blue] (box17) at (6,-11) { Special Purpose };
    \node[ text=blue] (box18) at (7,-2) {\begin{tabular}{c} Encoder \\ (Iterated \\ 6 times) \end{tabular}};

\end{tikzpicture}
 \label{fig:blocks}

\end{figure}

The successive neural layers implemented in Python TensorFlow and Keras are represented in Figure  \ref{fig:blocks} and their functioning is explained below. We explain in details below the functioning of each layer.

\section{Neural network model details} \label{sec:architecture} 
\subsection{Analysis of the different layers}
\begin{enumerate}

\item The \textbf{Input tensor} defines the shape of an input  sequence (instance). It appears as a  tensor of shape $(None,l,d)$ "None" being related to the batch size, which is not explicited in the input tensor construction.
\begin{lstlisting}[language=Python]
inputs=keras.Input(shape=(32,16))\end{lstlisting}

\item The \textbf{MultiHeadAttention Layer} produces output tensors of the same shape ($l$,$d$) as the inputs.
\begin{lstlisting}[language=Python]
x = layers.MultiHeadAttention(key_dim=head_size, num_heads=num_heads, dropout=dropout) (x, x)\end{lstlisting}

	\begin{itemize}
	\item This layer is explained in more details in the next section. It has two arguments. The first argument is the instance used to calculate the  Query and the second argument is the instance used to calculate the Key and the Value. In the Encoder part of a Transformer these two arguments are the same. The head\_size is the dimension chosen for the Queries, Keys and Values. In the original paper  "Attention is All You Need" by Vaswani and all, the dimension of the model is 512, the number of heads is 8 and the head\_size is 512/8=64. Here, we keep the 64 for the head\_size. For the dropout, we do not use the dropout layer optionality offered by the Keras MultiHead Attention Layer but, implement it separately. Therefore, we use "dropout=0" and add the following layer.

	\item A Dropout Layer
\begin{lstlisting}[language=Python]
x = layers.Dropout(dropout)(x)\end{lstlisting}
	The dropout layer has no learnable parameters and simply transforms any input instance of shape  ($l$,$d$) by replacing some units by zero in the  training phase (while keeping them identical when predicting). The purpose of the dropout layer is to create some robustness when learning the parameters of the model. We take dropout=0.25.
	\end{itemize}

\item An Additive  Layer
\begin{lstlisting}[language=Python]
res = x + inputs\end{lstlisting}
This layer has no learnable parameters and adds each unit of the dropout layer's output instance $ (xd_1,xd_2,\cdots, xd_l)$ to each unit of an input instance $(x_1,x_2,\cdots,x_l)$.
$$ \mbox{input} + \mbox{dropout layer's output} \longrightarrow  (xd_1+ x_1,xd_2+x_2,\cdots, xd_l+ x_l)$$ 
so, it produces a tensor of shape  (None, $l$,$d$).		
	
	There is no learnable parameter here.	

\item A Normalisation Layer (see Tensorflow \cite{TFNL})
\begin{lstlisting}[language=Python]
x = layers.LayerNormalization(axis=-1,epsilon=1e-6)(res)\end{lstlisting}
Layer normalisation is done independently for each sequence (contrarily to batch normalisation), by normalizing along a single axis or multiple axis. So, the mean and variance calculated along this/these axis become respectively  zero and 1. By default, Keras normalizes along the last axis. In a transformer the normalization is done along the feature axis (see  \cite{Hinton2016}, \cite{Lan2020}), i.e the dimension of the model. So, as the input tensor of the layer is of shape $(None,l,d)$ by saying "axis=-1" (or "axis=2") the normalisation is done in the dimension $d$ as usual.

 So, if $x_i \in \R^d$ is the ith element of a sequence (instance)  and $x_i^j$ is the value of its feature $j\in [1,d]$, $x_i$ is first transformed into 
 $$\tilde{x}_i=\frac{1}{\sigma(x_i)+ \epsilon}(x_i- \mu(x_i)1_d)$$ 
 
 with $\mu(x_i) =\frac{1}{d} \sum\limits_{j=1}^d x_i^j$ and  $\sigma(x_i) =\sqrt{\frac{1}{d} \sum\limits_{j=1}^d (x_i^j-\mu(x_i))^2}$ 
 
The (small) arbitrary parameter $\epsilon$ is chosen to insure that no division by zero occurs. So, each $\tilde{x}_i$ of a sequence gets normalized features in $\R^d$ of expectation $0$ and variance 1.
 
 The second thing that the normalization layer does is an affine transformation for the axis which has been normalized. 
 Here, the normalization occurs over [axis=-1] of shape ($d$) so, the shapes of the scale-tensor $\gamma$ and center-tensor $\beta$ are ($d$). The $\tilde{x}_i^j$ are transformed by these learnable tensors $\beta=(\beta_1,\beta_2, \cdots \beta_d )$ and $\gamma=(\gamma_1,\gamma_2, \cdots\gamma_d )$  in the following way.

$$ \tilde{x}_i^j \longrightarrow \gamma_j \tilde{x}_i^j + \beta_j $$
The number of learnable parameters is $2d$. 

Then comes the Position-wise Feed Forward block with:

\item A 1D Convolution Layer (see Keras  \cite{TFConv1D}). 
\begin{lstlisting}[language=Python]
x = layers.Conv1D(filters=ff_dim, kernel_size=1, activation="relu")(x)\end{lstlisting}	
Each filter corresponds to a transformation which is affine if "use bias= True" (which is the default) and otherwise linear.
Here, the shape of the input tensor is $(None,l,d)$ and each affine transformation is done along "axis=1". 
As "kernel size=1", the transformation is done one $x_i$ at a time.
$$x_i \longrightarrow w_i^{\prime}x_i +b_i $$ with $w_i$ and $b_i$  learnable of shape $(l)$.
The output tensor is of shape (None, $l$, ff\_dim). 
In Vaswani \cite{Vaswani2023} ff\_dim $= 4\times d$ and the activation function is "relu". We use here the same multiplier $4$, and the same activation function. The number of learnable parameters is $(d+1)\times l \times $ ff\_dim.

\item A Dropout Layer
\begin{lstlisting}[language=Python]
x = layers.Dropout(dropout)(x)\end{lstlisting}
\item Another  1D Convolution Layer, but this time with a number of filters equal to $d$ to produce an output of shape ($l$,$d$).
\begin{lstlisting}[language=Python]
x = layers.Conv1D(filters=inputs.shape[-1], kernel_size=1)(x)\end{lstlisting}	

This layer finishes the Position-wise Feed Forward block. Then comes,

\item An Additive  Layer
\begin{lstlisting}[language=Python]
res = x + inputs\end{lstlisting}
\item A Normalization Layer
\begin{lstlisting}[language=Python]
x = layers.LayerNormalization(axis=-1,epsilon=1e-6)(inputs)\end{lstlisting}	

This Encoder Block is iterated 6 times in Vaswani  \cite{Vaswani2023}. This is also what we do in our base case by defining 
"num transformer blocks=6". 

Note that our program performs better when the last Normalization Layer 9 is moved in front of the Multihead Layer and it is what we do in our base case as indicated by the red arrow in Figure  \ref{fig:blocks}.

The last part of the model 	is specific to our classification objective and its input tensor of shape (None,$l$,$d$) must be transformed into a tensor of shape (None, $l$,$k$). This last block is made of the following layers.

\item A GlobalAveragePooling1D layer. \\  
\begin{lstlisting}[language=Python]
x = layers.GlobalAveragePooling1D(data_format="channels_first")(x)\end{lstlisting}
By default the average is done along the sequence axis, transforming an input tensor of shape $(None,l,d)$ into an output tensor of shape $(None,d)$.
Here, the model performs better when the average is done along the features (channels) axis.
By adding the argument ''data\_format="channels\_first"' the layer treat the input tensor as if its shape was $(None, d,l)$ therefore calculating the averages along the last axis. Therefore, with this argument in our model the output tensor has the shape $(None,l)$.

\item A Dense Layer with a number of neurons equal to "mlp units".
We take in our base case  "mlp units =[10]" and  implement a single layer with no iteration. To implement a succession of $p$ Dense Layers, the number of units of these successive layers are entered as a numpy array "mlp units = $[n_1, n_2,....n_p]"$.
\begin{lstlisting}[language=Python]
x = layers.Dense(dim, activation="relu")(x)\end{lstlisting}

\item A Dropout Layer 
\begin{lstlisting}[language=Python]
x = layers.Dropout(mlp_dropout)(x)\end{lstlisting}
in our base case we take  mlp\_dropout$=0.25$.
\item A final Dense Layer which produces an output of shape $($n\_classes,$)$ i.e $(7,)$.
\begin{lstlisting}[language=Python]
outputs = layers.Dense(n_classes, activation="softmax")(x) \end{lstlisting}
	

\end{enumerate}

The model is implemented with Python tensorflows on Google Colab (Python torch is also popular alternative).

\subsection{The Multi Head Transformer-Encoder}

The different layers of the MultiHead Transformer-Encoder are described in more details here:
 \begin{enumerate}

\item a MultiHead Attention Layer. This block produces output tensors of the same shape as the input tensors.
$$(None, l,  d) \longrightarrow (None, l,  d) $$
The operations conducted are as follows
\begin{itemize}
\item each Head transforms an input tensor of shape $(None, l,d)$  into a tensor of shape $(None,l,d_v)$ via the Attention Mechanism
\item these $nb_{heads}$ tensors are then concatenated to form a tensor of shape $(None, l,nb_{heads} \times d_v) $
\item then a linear layer transforms this tensor into a tensor of shape $(None, l, d)$
\end{itemize}

The MultiHead Attention Mechanism works as follows:\\
For each "Head", each observation $x_i$ of shape $(d,)$ is transformed linearly according to some learnable matrices \\
$W_K$ of dim $(d,d_k)$ to produce a Key-tensor $x_i^K=  x_i W_K$ of shape $(d_k,)$ \\
$W_Q$ of dim $(d,d_k)$ to produce a Query-tensor $x_i^Q=  x_i W_Q$  of shape $(d_k,)$\\
$W_V$ of dim $(d,d_v)$ to produce a Value-tensor $x_i^V= x_i W_V $ of shape $(d_v,)$.\\

We call
$K =\begin{pmatrix}
  x_1^K \\
  \vdots \\
  x_l^K  \\
\end{pmatrix}$,
$Q =\begin{pmatrix}
  x_1^Q \\
  \vdots \\
  x_l^Q  \\
\end{pmatrix}$, 
$V =\begin{pmatrix}
  x_1^V \\
  \vdots \\
  x_l^V  \\
\end{pmatrix}$

These matrices are of respective dimensions $(l,d_k)$, $(l,d_k)$, $(l,d_v)$.

We call Attention Matrix $A$ the matrix of dimension $(l,l)$ defined as: \\
$$ A = softMax\Big(\frac{1}{\sqrt{k}}QK^{\perp}\Big)$$ 
The softmax function is applied line by line so that the sum of each line of $A$ is one and then we calculate the quantity
$$ AV$$
Thus, we obtain for each Head a matrix $AV $ of dimension  $(l,d_v)$  which are concatenated to form a matrix 
$\begin{pmatrix}
  z_1\\
  \vdots \\
  z_l \\
\end{pmatrix}$ of dimension $(l,nb_{heads} \times d_v)$.

Last step  is to use a learnable matrix $W_O$ of shape $(nb_{heads} \times d_v,d)$ to transform $z$ into 
$V =\begin{pmatrix}
  z_1W_O \\
  \vdots \\
  z_lW_O \\
\end{pmatrix}$ of shape $(l,d)$.

 \end{enumerate}

\subsection{Performance of the model}

 \subsubsection{Crossentropy}
The cross entropy between a target probability $P=(p_1,p_2,\cdots, p_k)$ of belonging to  the $k$ buckets and a softmax prediction $Q=(q_1,q_2,\cdots, q_k)$  is defined as:

$$ H(P,Q)=-\sum\limits_{j=1}^k p_j ln(q_j) \geq H(P,P)$$

and  $\min\limits_{Q} H(P,Q)$ is reached for $Q=P$.

In the learning phase, the cross entropy is minimised between the dirac probabilities $P_i=(1_{B_1}(y_{i+l}),1_{B_2}(y_{i+l}),\cdots, 1_{B_k}(y_{i+l}) )$ observed and the probabilities $Q_i=(q^i_1,q^i_2,\cdots, q^i_k)$  predicted  from the observation $[x]_i$. For a batch, made of $\# B$ sequences, the loss function minimised is : 

$$\frac{1}{\# B} \sum\limits_{i \in B} H(P_i,Q_i)$$

With simulated data we can calculate the target probabilities $T_i$ based on all events (hidden and observable) occured up to time $i+l-1$  as
$$t^i_j=P(y_{i+l} \in B_j | H_{i+l-1}=h_{i+l-1})$$

\subsubsection{Categorical accuracy}
The categorical accuracy provides another measure of performance by measuring the percentage of correct bucket predictions (the bucket predicted for $y_{i+1}$ is the one maximising the $q^i_j$ for $j \in \llbracket 1,k\rrbracket$).
 
There are $k$ buckets. At the beginning, the model does not know how to use the $[x]_i$ to predict and makes prediction at random and therefore predicts correctly in $\frac{1}{k}$ of the cases and we observe and accuracy of $\frac{1}{k}= \frac{1}{7}= 14.28\%$. 

 After training with the simulated data, the accuracy approaches $30\%$, which is not far from the best possible accuracy derived with the $T_i$. 
  
 Indeed, the best possible accuracy for bucket prediction is reached when the target probabilities $T_i$ are used and when the predicted bucket is defined as  $$ \arg\max\limits_{j} t_j^i $$
 with this method, the Categorical Accuracy reached from the sample of 24131 observations is $31.85\%$ on the train set and $32.00\%$ on the test set.

\subsubsection{Pointwise analysis of the predictions } 
With simulated data, we can calculate the target probabilities $T_i$ based on all variables (hidden and observable) occured up to time $i+l-1$  as
$$t^i_j=P(y_{i+l} \in B_j | H_{i+l-1}=h_{i+l-1})$$ that the classifier should ideally be able to approximate, and compare the quantities  $H(T_i,Q_i) $ to our targets $H(T_i,T_i) $. The comparison is done by calculating the average of these two quantities on the train set and on the test set.

When the model starts training it does not know how to use past observations and therefore find a probability $Q$ independent of the $[x]_i$ which solves 
$$ \min\limits_{Q} -\sum\limits_{i=1}^k \frac{1}{k}ln(q_i) $$ Therefore, the  solution is,  $\forall i \in \llbracket 1,k \rrbracket, q_i=  \frac{1}{k}= \frac{1}{7}$ which results in a loss function equal to $ln(k)=ln(7)=1.9459$, which is what we get at the beginning of the optimisation process. As the optimisation progress, we expect the loss function to get closer to 
$$\underset{i}{Average}  \{ H(T_i,T_i)\} \sim 1.63 $$

and we manage to get pretty close.

\begin{figure}[htbp!]
\centering
\includegraphics[width=1.\textwidth]{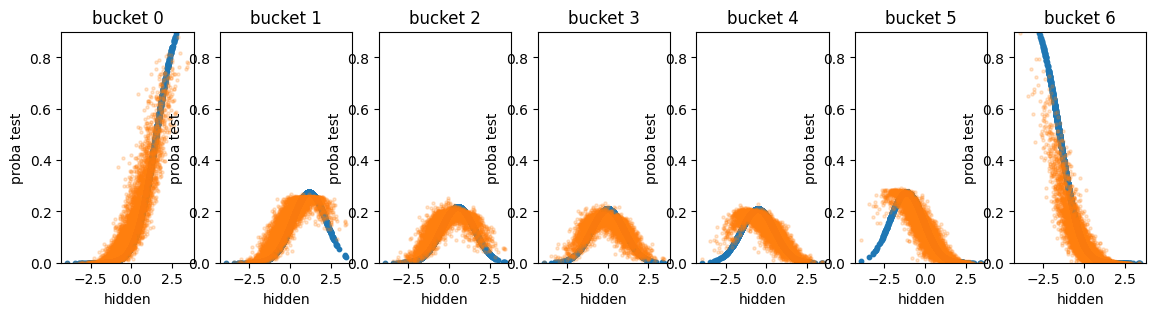}
\caption{Predictions for the synthetic 
	stochastic process~\eqref{eq:ouprocess} with 24131 observations, 30 epochs.}
\label{fig:example1}
\end{figure}

\begin{figure}[htbp!]
  \centering
  \includegraphics[width=1.\textwidth]{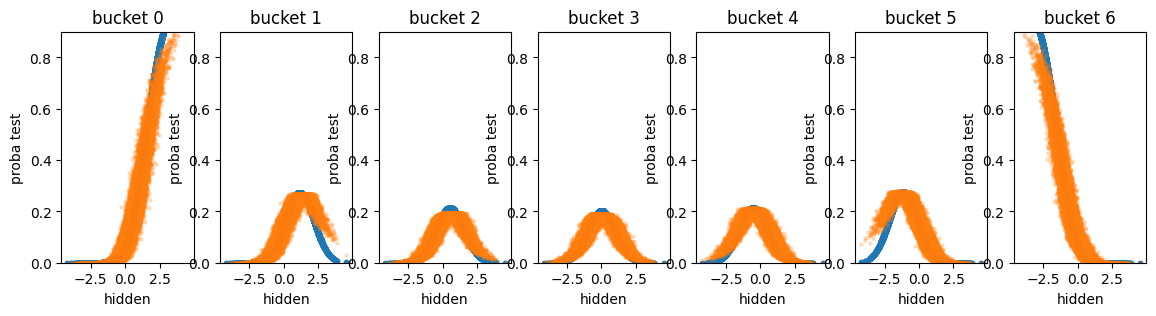}
  \caption{Predictions for the synthetic 
  	stochastic process~\eqref{eq:ouprocess} with  241310 observations, 40 epochs.}
  \label{fig:example2}
\end{figure}

For each $[x]_i$ we compare the probabilities $Q_i$ predicted by the model to the target probabilities  $T_i$ by plotting for each bucket $j \in \llbracket  1,k \rrbracket$ the points $(h_{i+l-1},q^i_j)$ and the points $(h_{i+l-1},t^i_j)$, see Figure \ref{fig:example1} and Figure \ref{fig:example2}.

\subsection{Parameters of the model} 

We make the arbitrary choice in our base model to work with sequences of length 32.  
The big difference between NLP and times series is that for time series there is no standard method to embed numbers. We choose here an arbitrary function $\phi$ for embedding. 
Note that, when we calculate in the Multihead Attention block the scalar products $\langle \phi(x_i) ,\phi(x_j)\rangle $ the choice of $\phi$ creates some Kernel $K(x_i,x_j)=\langle \phi(x_i) ,\phi(x_j)\rangle $ which generally speaking are important tools in classification and prediction problems.

\begin{table}[h]
    \centering
    \begin{tabular}{|>{\centering\arraybackslash}m{6cm}|>{\centering\arraybackslash}m{5cm}|}
        \hline
        Base case NLP encoder  & Base case time series  prediction \\ 
        \hline
        length sequence: $l \sim 1024$  		& $l=32$ \\ 
        \hline
        embedding dimension:  $d=512 $ 		& $d=\frac{l}{2}=16$ \\  
         \hline
       Positional Encoding:  sinus and cosinus 			& sinus and cosinus \\  
         \hline
        number of heads: $h =8$  				&   $h=8$   \\
         \hline
        head size:  $d_k =\frac{d}{N_h}=64$ 		& $d_k = 64$  \\
         \hline
        number of block iterations: $N=6$  		& $N=6$ \\
         \hline
        number of units FFN: $d_{ff}= 4\times d = 2048 $ 	&  $d_{ff}= 4\times 16= 64 $  \\
        \hline \hline
          									& Dense layer classifier  $mlp\_units = [10]$  \\
	\hline
    \end{tabular}
    \caption{Parameters of the model.}
    \label{tab:exemple1}
\end{table}

The model uses, the categorical cross entropy for the loss function, Adam for the optimizer and the "categorical accuracy" for the metrics.

\begin{lstlisting}[language=Python]
model.compile( loss="categorical_crossentropy",
optimizer=keras.optimizers.Adam(learning_rate=1e-3),
metrics=["categorical_accuracy"])\end{lstlisting}

We use 80\% of the times series for learning and 20\% for test. In the learning phase the model uses 80\% of the learning sequences for calibration and 20\% for validation.

\section{Results} \label{sec:results}

\subsection{Prediction mean reverting Ornstein-Uhlenbeck synthetic data}

We present here some results obtained in our base model; the parameters, inspired by the base model of Vaswani \& Al  \cite{Vaswani2023}, are given in Table \ref{tab:exemple1}.

We considered here a trajectory of the synthetic 
stochastic process~\eqref{eq:ouprocess} with
$24131$ data points, the same as the number of daily data we will use for the prediction on the S\&P500 market index. The results are given in Table \ref{tab:example2}.
\begin{table}[h]
    \centering
    \begin{tabular}{|p{1.1cm}|p{1.7cm}|p{1.9cm}|p{1.9cm}|p{1.5cm}|p{1.5cm}|}
        \hline
        Number of epochs & Number of observations & Train~set: Loss~$H(P,Q)$, Accuracy&
         Test~set: Loss~$H(P,Q)$, Accuracy	& Train~set $H(P,T)$, $H(T,T)$ 	& Test~set  $H(P,T)$, $H(T,T)$  \\
	\hline
         30                       & 	24131	 			& 1,681		         & 1.697			& 1.626			 & 1.636     \\
            		      & 	  		         	 	& 30.33\%		 	& 28.66\%			& 1.630			 & 1.628 \\
        \hline
        30			 & 241310			 & 1,656			 & 1.656			 & 1.631			 & 1.630 \\
        				 & 				 & 30.60\%		 & 30.74\%			 & 1.631			 & 1.629 \\
          \hline
        40			 & 241310			 & 1,657			 & 1.657			 & 1.631			 & 1.630 \\
        				 & 				 & 30.56\%			 & 30.77\%			 & 1.631			 & 1.629 \\       
        \hline
    \end{tabular}
    \caption{Analysis of the results obtained with simulated data.}
    \label{tab:example2}
\end{table}

We se that the model enables accurate prediction of the bucket probabilities (the blue points correspond to the perfect probability predictions with the $T_i$) but a significant number of instances are needed to train it.

\subsection{Prediction on the S\&P500 Index}

\begin{figure}[htbp!]
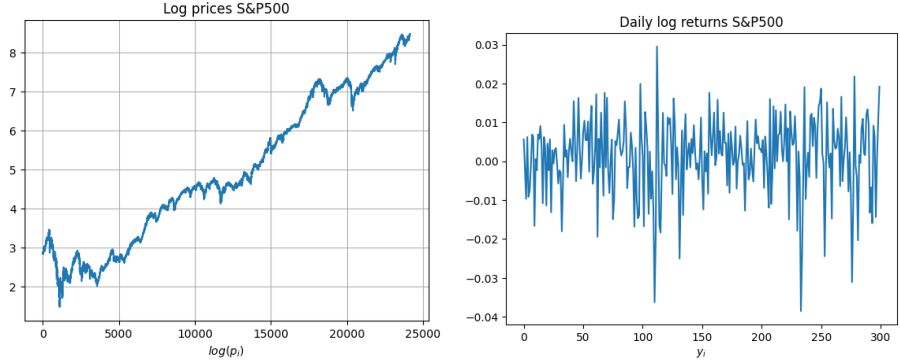

  \centering
  \includegraphics[width=0.49\textwidth]{SP_Log_process.pdf}
  \includegraphics[width=0.49\textwidth]{SP_daily_returns.pdf}
  \caption{{\bf Left:} Log prices of the S\&P500 over the period.
  	{\bf Right:} first 300 daily log returns $y_1, ...,y_{300}$.}
  \label{fig:trajectory2}
\end{figure}

We now make predictions on market data using the closing prices for the  S\&P500 Index from  the $30^{th}$ December 1927  to the $1^{th}$ February 2024.
We dispose of 24131 price observations.
Here, the variables $h_i$ correspond to the logarithm $ln(p_i)$ of the prices observed  and we build predictions for the 
\begin{equation} y_i = ln(p_i)- ln(p_{i-1}) \end{equation} which are the daily log returns observed (see Figure  \ref{fig:trajectory2}).

In the first program we predict the bucket for  $y_i$, i.e, we try to guess the probability distribution of the daily log return for the next business day.
In the second program we predict the bucket for $y_{i}^2$ (the quadratic variation of the log prices) for the next business day.

\subsection{Prediction of the next return $y_i$} 

\begin{figure}
\centering
\includegraphics[width=1.\textwidth]{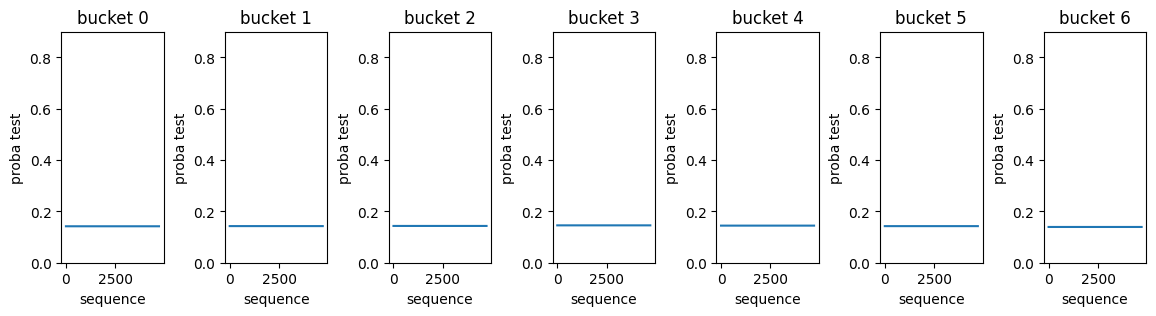}
\caption{bucket predictions for $y_{i+l}$ on the test set for the S\&P500.}
\label{fig:example5}
\end{figure}

When we  applied the base model it became stuck in predicting constant bucket probabilities, without differentiating between the $[x]_i$ (see Figure \ref{fig:example5}).
We used $10$ epochs and for all $i$ in the train set the probability predictions for the $k=7$ buckets (defined by the 
quantiles $a_j$, $j=1,...,k-1$ as described earlier) were  all equal  to :
$$\{0.14200473, 0.14268069,  0.14331234, 0.1455955, 0.1445429, 0.1425599, 0.1393042\}$$
As all are equal, this means that the model is failing to make an interesting prediction because it fails to take into account the conditional information $[x]_i$.

We did not spend too much efforts trying to improve these predictions, for which we had limited hopes, and directed our efforts towards the $y_i^2$ prediction objective.

\subsection{Prediction of the next quadratic variation $y_{i+l}^2$}

Here, the aim is from an observed $[x]_i$, to predict the bucket for $y_{i+l}^2$.  This is the first step to predict the gamma cost of hedging options and future volatility. As statistical studies using GARCH models for example have advocated for some degree of predictibility, we expect that our predictions will  surpass random chance in this case.

The model parameters remain the same as before and we continue to refrain from  positional encoding, which appears  to hinder learning. Fifty epochs seem to be enough to calibrate and we get the following results.  

\begin{table}[h]
    \centering
    \begin{tabular}{|p{1.7cm}|p{2cm}|p{2.2cm}|p{2.2cm}|}
        \hline
        Number of 	     & Number of  			& Train set			& Test set       		   \\
       	epochs	      & observations    		&  Loss $H(P,Q)$		& Loss $H(P,Q)$	 	     \\
			     &					& Accuracy	         		& Accuracy	          		     \\
	\hline
         50                       & 	24131	 			& 1,861		         & 1.876			    \\
            		      & 	  		         	 	& 21.92\%		 	& 22.84\%			 \\
     
        \hline
    \end{tabular}
    \caption{bucket predictions  for $y_i^2$.}
    \label{tab:example}
\end{table}

\begin{figure}
  \centering
  \includegraphics[width=1.\textwidth]{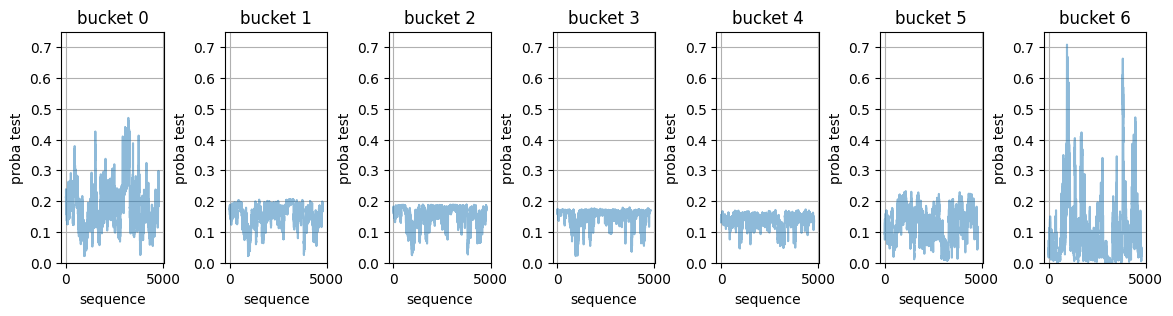}
  \caption{bucket predictions for $y_{i+l}^2$ on the test set for the S\&P500.}
  \label{fig:quadratic}
\end{figure}

The predicted probabilities for the $k=7$ buckets, corresponding to the successive sequences $[x]_i$ in the test set, are illustrated in Figure \ref{fig:quadratic}. While the results are imperfect, they surpass those of a random prediction across the $k$ buckets (which are equally likely in the training set), yielding a cross-entropy of $\ln(k)=\ln(7)=1.9459$ and an accuracy of $\frac{1}{k}=\frac{1}{7}=14.28\%$.

\begin{figure}
  \centering
  \includegraphics[width=1.\textwidth]{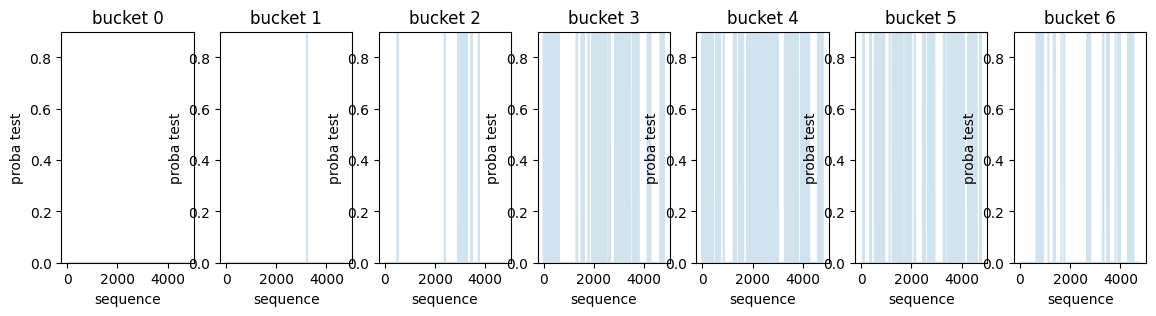}
  \caption{Naive classifier for $y_{i+l}^2$  on the test set for the S\&P500.}
  \label{fig:naive}
\end{figure}

Another natural benchmark for evaluating our classifier is the "naive" classifier, which assigns $y_{i+l}^2$ (with 100\% probability) to the same bucket as $\frac{1}{l} \sum\limits_{j=i}^{j=i+l-1}y_j^2$. For this classifier, the accuracy on the test set is $19.27\%$, and the predicted classes for the successive $[x]_i$ are depicted in Figure \ref{fig:naive}. Once again, our classifier outperforms this naive approach, which is an encouraging outcome.

Undoubtedly, there is room for further refinement to enhance the performance of the Encoder Classifier. This could involve adjustments to the structure, choice of encoding method, model dimensionality, inclusion of additional variables, and so forth. Nonetheless, we view this current work as a promising starting point, yielding some encouraging results.

\section{Conclusion} \label{sec:conclusion}.

In this study, we applied a transformer encoder architecture to times series prediction. The significant differences from LLM applications of transformers is that here we are dealing with numerical data (of dimension 1) and not tokens  embedded in spaces of dimension 512 or 1024 (with a specific logic behind the embedding process). We felt compelled to embed the numbers into a higher dimension space  in order to prevent  significant information loss in the normalization layers (that encoders possess). That being said, our approach to embedding numbers is rather naive and just based on the premise that by embedding numbers with a function $\phi$ the scalar product we create in the process $\langle \phi(x_i), \phi(x_j)\rangle$ may be related to some useful Kernel $K(x_i,x_j)= \langle \phi(x_i), \phi(x_j)\rangle$. We did not have to do too much fine tuning to have a working model in the simulation case, whose results were therefore  encouraging. However, we were surprised to find that positional encoding did not  enhance prediction accuracy or improve convergence speed. Additionally, we noticed that altering the value of $\theta$ in the simulations could degrade model performance or impede learning, necessitating adjustments. The main differences between our structure and the standard encoder is the normalization layer that we put before the multi-head instead of before the classifier block. This deviation was motivated by the recommendation against normalizing before a dense classification layer. Regarding the prediction of the square of the daily returns of the S\&P500, which can be useful in terms of volatility and cost of gamma hedging predictions, we find the results encouraging. However, we believe that the model's performance could be further improved through more fine-tuning of the transformer's structure, as well as the embedding process and selection of additional observation variables.


\begin{thebibliography}{9}

\bibitem{Zeng} Ailing Zeng, Muxi Chen, Lei Zhang,  Qiang Xu.
 (2023) \emph{Are Transformers Effective for Times Series Forecasting?} The Thirty-Seventh AAAI Conference on Artificial Intelligence, 2023 (AAAI-23).

\bibitem{Vaswani2023}
Ashish Vaswani, Noam Shazeer, Niki Parmar, Jakob Uszkoreit, Llion Jones, Aidan N. Gomez, Lukasz Kaiser, Illia Polosukhin,  (2017) \emph{Attention Is All You Need}, arXiv:1706.03762v7, 2 August 2023.

\bibitem{TFConv1D} Conv1D layer,
 \url{https://keras.io/api/layers/convolution_layers/convolution1d/}, retrieved Feb. 29th 2024

\bibitem{Eli}
Eli Simhayev, Kashif Rasul, Niels Rogge,  (2023) \emph{Yes, Transformers are Effective for Time Series Forecasting (+ Autoformer)}, \url{https://huggingface.co/blog/}, 16 June 2023.

\bibitem{Hinton2016}
Jimmy Lei Ba, Jamie Ryan Kiros, Geoffrey E.Hinton, (2016) \emph{Layer Normalization}, arXiv:1607.06450v1, 21 July 2016.


\bibitem{Wen} Qingsong Wen, Tian Zu, Chaoli Zhang, Weiqi Chen, Ziqing Ma, Junchi Yan, Liang Sun.
 (2023) \emph{Transformers in Times Series: A Survey}, Proceedings of the Thirty-Second International Joint Conference on Artificial Intelligence (IJCAI-23).


\bibitem{Lan2020}
Ruibin Xiong,  Yunchang Yang, Di He,  Kai Zheng,  Shuxin Zheng,   Chen Xing, Huishuai Zhang,  Yanyan Lan,  Liwei Wang,  Tie-Yan Liu,  (2020) \emph{On Layer Normalization in the Transformer Architecture}, arXiv:2002.04745v2, 29 June 2020.



\bibitem{TFNL} TensorFlow v2.15.0.post1, (2024) \url{https://www.tensorflow.org/api_docs/python/tf/keras/layers/LayerNormalization}, last updated 2024-01-11 UTC.








\end{thebibliography}
\end{document}